\documentclass[conference]{IEEEtran}
\IEEEoverridecommandlockouts
\usepackage{cite}
\usepackage{amsmath,amssymb,amsfonts}
\usepackage{algorithmic}
\usepackage{graphicx}
\usepackage{textcomp}
\usepackage{xcolor}
\def\BibTeX{{\rm B\kern-.05em{\sc i\kern-.025em b}\kern-.08em
    T\kern-.1667em\lower.7ex\hbox{E}\kern-.125emX}}
\usepackage{bm}   % 放在 amsmath 之后也可以    
\usepackage{amsmath}
\usepackage{graphicx}
\usepackage[table,xcdraw]{xcolor}
\usepackage{diagbox}
\usepackage{multirow}
\usepackage{tabularx}
\usepackage{xcolor}
\newcolumntype{Y}{>{\centering\arraybackslash}X}
\usepackage{amsfonts,amssymb}

\def\BibTeX{{\rm B\kern-.05em{\sc i\kern-.025em b}\kern-.08em
    T\kern-.1667em\lower.7ex\hbox{E}\kern-.125emX}}
\usepackage[font=small,labelfont=bf,tableposition=top]{caption}

\usepackage{blindtext}
\title{EFDiT: Efficient Fine-grained Image Generation Using Diffusion Transformer Models}

\author{
\IEEEauthorblockN{Kun Wang\textsuperscript{1, 2} Donglin Di\textsuperscript{3} Tonghua Su\textsuperscript{1,2 $^{\dagger}$} and Lei Fan\textsuperscript{4$^{\dagger}$}}\thanks{$^{\dagger}$Corresponding author.}
\IEEEauthorblockA{
\textsuperscript{1}Harbin Institute of Technology, Harbin, China\\
\textsuperscript{2}Chongqing Research Institute of HIT, Chongqing, China\\
\textsuperscript{3}Li Auto, Beijing, China\\
\textsuperscript{4}University of New South Wales, Sydney, Australia }
\IEEEauthorblockA{
wk1360178179@gmail.com, donglin.ddl@gmail.com, thsu@hit.edu.cn, lei.fan1@unsw.edu.au}
}

\let\oldtwocolumn\twocolumn
\renewcommand\twocolumn[1][]{%
    \oldtwocolumn[{#1}{
    \begin{center}
           \includegraphics[width=0.9\textwidth]{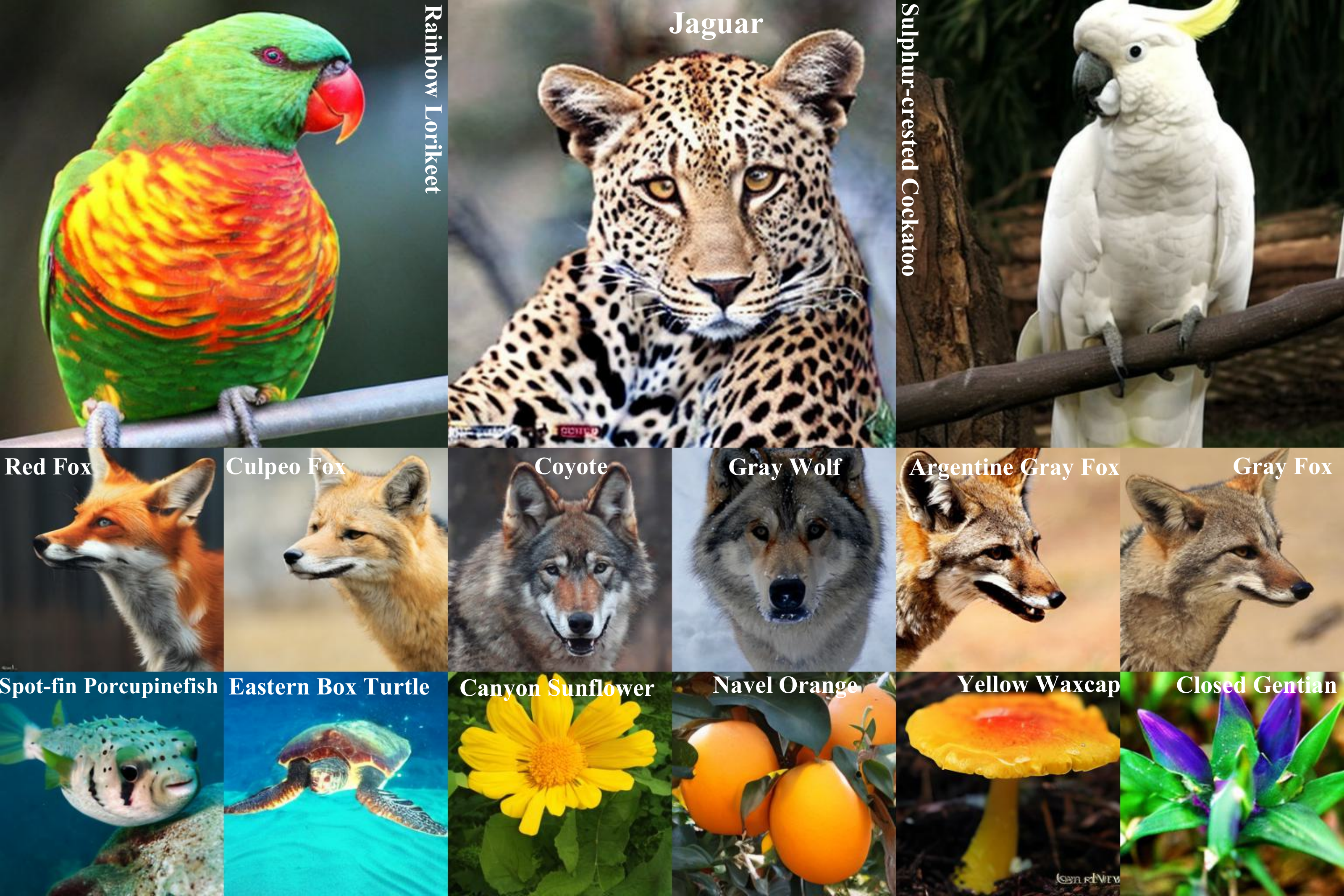}
           \captionof{figure}{The figure presents examples of fine-grained images generated by the EFDiT model, where the second row shows fine-grained images of similar categories, and the third row displays high-quality fine-grained images.}
           \label{fig:fig1}
        \end{center}
    }]
}

\begin{document}

\maketitle

\begin{abstract}
Diffusion models are highly regarded for their controllability and the diversity of images they generate. However, class-conditional generation methods based on diffusion models often focus on more common categories. In large-scale fine-grained image generation, issues of semantic information entanglement and insufficient detail in the generated images still persist. This paper attempts to introduce a concept of a “tiered embedder" in fine-grained image generation, which integrates semantic information from both super and child classes, allowing the diffusion model to better incorporate semantic information and address the issue of semantic entanglement. To address the issue of insufficient detail in fine-grained images, we introduce the concept of super-resolution during the perceptual information generation stage, enhancing the detailed features of fine-grained images through enhancement and degradation models. Furthermore, we propose an efficient ProAttention mechanism that can be effectively implemented in the diffusion model. We evaluate our method through extensive experiments on public benchmarks, demonstrating that our approach outperforms other state-of-the-art fine-tuning methods in terms of performance.  
\end{abstract}

% \begin{abstract}
%     Diffusion models are highly regarded for their controllability and the diversity of images they generate. However, class-conditional generation methods based on diffusion models often focus on more common categories. In large-scale fine-grained image generation, issues of semantic information entanglement and insufficient detail in the generated images still persist. This paper attempts to introduce the concept of a “tiered embedder" in fine-grained image generation, which integrates semantic information from both superclasses and subclasses, allowing the diffusion model to better incorporate semantic information and address the issue of semantic entanglement. To address the issue of insufficient detail in fine-grained images, we introduce the concept of super-resolution during the perceptual information generation stage, enhancing the detailed features of fine-grained images through enhancement and degradation models. Furthermore, we propose an efficient ProAttention mechanism that can be effectively implemented in the diffusion model. We evaluate our method through extensive experiments on public benchmarks, demonstrating that our approach outperforms other state-of-the-art fine-tuning methods in terms of performance.
% \end{abstract}

\section{Introduction}
Fine-grained image generation \cite{peng2024frih} focuses on synthesizing images that accurately capture detailed characteristics across a wide range of categories \cite{fan2022grainspace,fan2025grainbrain}, each containing multiple subcategories with subtle yet significant variations. With deep learning techniques~\cite{fan2023av4gainsp,fan2024patch}, early studies predominantly relied on Generative Adversarial Networks (GANs)~\cite{goodfellow2014generative,fan2022fast}, VAEs \cite{kingma2013auto}, and Flow-based models\cite{kingma2018glow}; however, these methods often faced challenges in producing high-quality, condition-aligned fine-grained images effectively. Recently, diffusion-based models \cite{wang2024towards,sun2024eggen} have emerged as the dominant approach in image generation due to their capability to model complex data distributions and generate high-resolution images with fine details. For instance, the Diffusion Transformer (DiT) \cite{peebles2023scalable} can generate high-quality images conditioned on inputs such as labels, enabling precise control over image attributes and ensuring alignment with specified conditions.

% In fine-grained image generation, some studies \cite{zaken2021bitfit, fan2022fridofeaturepyramiddiffusion} focused on fine-tuning pretrained transformer models, typically by directly extracting semantic information from subcategories and embedding it into the diffusion model to generate the corresponding fine-grained images. For example, classifier-free guidance \cite{ho2022classifier} method jointly trains conditional and unconditional diffusion models \cite{kim2023dcface, wang2023styleinv, chae2024semantic}. However, in category-conditioned fine-grained image generation, while these models generate images that broadly reflect the general semantics of given labels (\textit{e.g.}, `bird'), they often struggle to accurately produce specific distinguishing features of closely related subcategories, such as particular bird species (\textit{e.g.}, `eagle') \cite{xie2023difffit}. This challenge arises due to the large number of categories and the visual similarity among related subcategories, and these models fall short of comprehending the intricate relationships between parent categories and their subcategories. 

In fine-grained image generation, some studies \cite{zaken2021bitfit, fan2022fridofeaturepyramiddiffusion} focused on fine-tuning pretrained transformer models, typically by directly extracting semantic information from subcategories and embedding it into the diffusion model to generate the corresponding fine-grained images. For example, classifier-free guidance \cite{ho2022classifier} method jointly trains conditional and unconditional diffusion models. However, in category-conditioned fine-grained image generation, while these models generate images that broadly reflect the general semantics of given labels (\textit{e.g.}, `bird'), they often struggle to accurately produce specific distinguishing features of closely related subcategories, such as particular bird species (\textit{e.g.}, `eagle') \cite{xie2023difffit}. This challenge arises due to the large number of categories and the visual similarity among related subcategories \cite{fan2023identifying}, and these models fall short of comprehending the intricate relationships between parent categories and their subcategories.

On the other hand, to reduce the high training costs, some studies \cite{chen2024pixart, yang2024structure} have explored fine-grained image generation by fine-tuning DiT models to adapt to new fine-grained image datasets. However, these methods typically generate fine-grained images by directly fine-tuning model parameters, resulting in a lack of texture details in the generated fine-grained images. However, we aim to continuously enhance texture details during the fine-grained image generation process to preserve the unique characteristics of the details. For images, different frequency components in the frequency domain correspond to their respective features, which provides strong theoretical support for realizing this idea.

% On the other hand, to reduce the high training costs, some studies \textcolor{red}{\cite{chen2024pixart, yang2024structure, liu2024efficient}} have explored fine-grained image generation by fine-tuning DiT models to adapt to new, fine-grained image datasets.
% However, these approaches often resulted in images with insufficient details \textcolor{red}{\cite{yu2023osrt}}. To gain a deeper understanding of this limitation, we analyzed the diffusion model and observed that the noise removal at the initial stages of the denoising process primarily affects low-frequency information, with a gradual impact on high-frequency details as the process progresses.
%`````'''''

To tackle the above challenges, in this paper, we divided the denoising process into two stages: semantic generation and perceptual generation responsible for the overall structure and local details. A degradation model is used to simulate the detail loss during the generation of fine-grained images in the diffusion model, while an enhancement model is utilized to reinforce the contours and fine details. Furthermore, we propose a tiered embedder designed to enrich the semantic information by incorporating superclass labels into the diffusion model. This embedder comprises both superclass and subclass encoders, which, during fine-tuning, adjust the bias and regularization layers to achieve a more comprehensive encoding of the hierarchical label structure. To further enhance the efficiency of fine-tuning the DiT model, we introduced a ProAttention, which reduces the time complexity of the attention mechanism to $L \log(L)$, thereby significantly improving the model’s inference speed. 

\begin{figure*}[!t]\centering
	\includegraphics[width=\linewidth]{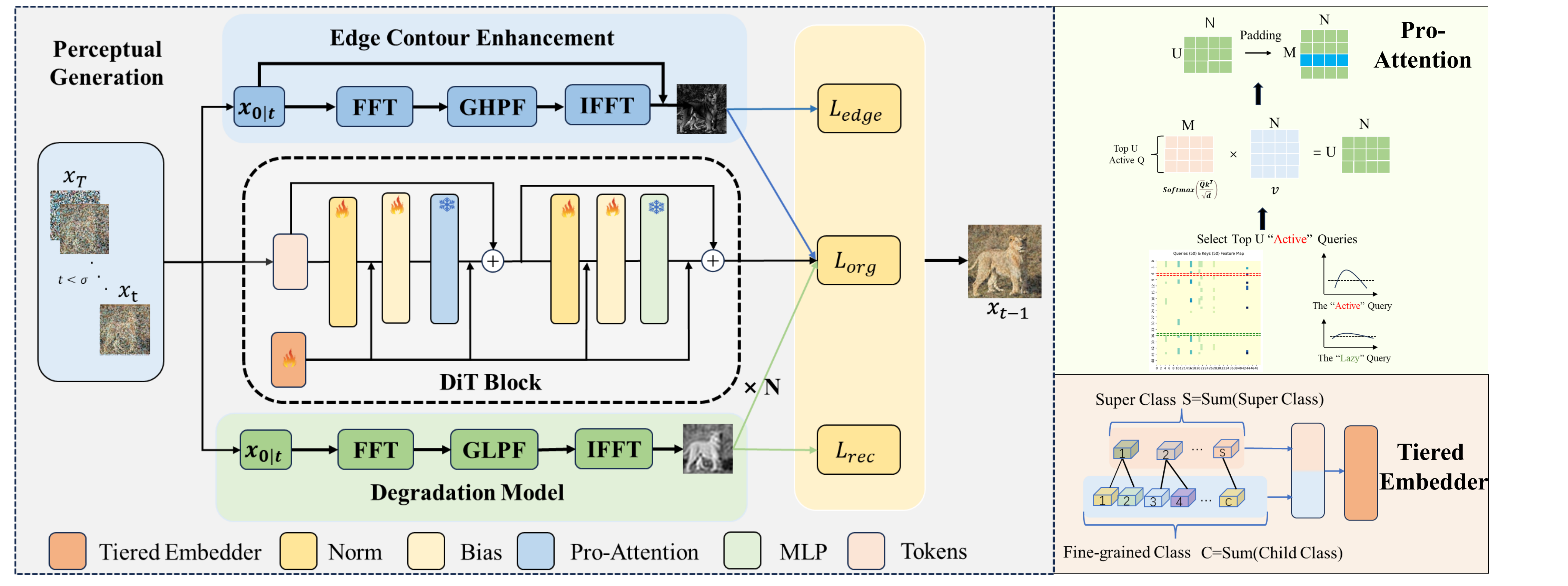}
	\caption{High-resolution fine-grained image generation architecture diagram. The model incorporates the “Tiered Embedder" shown in the bottom right to introduce superclass information into the model, introduces the ProAttention mechanism to enhance training efficiency, and integrates the concept of super-resolution during the denoising process to generate high-quality fine-grained image pixels.}
    \label{fig:struction}
\end{figure*}

% In our analysis of the DiT model, we identified a long-tail phenomenon in the attention mechanism, as shown in Figure \ref{fig_long_tail}. To address this issue, we introduced the ProAttention mechanism, which reduces the time complexity of the attention mechanism to $Llog(L)$, thereby improving the model's inference speed. This innovation not only effectively tackles the challenges posed by high pixel demands in fine-grained image generation but also enhances the model's efficiency in processing large-scale datasets, ensuring the generation of high-quality images.

Thus, we summarize our contributions as follows: 
\begin{itemize}
    \item  We innovatively propose the concept of superclasses, enhancing the semantic understanding capabilities of diffusion models in fine-grained image generation and offering a new perspective for generation strategies.
    \item We present super-resolution into diffusion models, enhancing the generation of fine-grained image details through degradation and enhancement models during the perceptual information generation stage.
    \item The introduction of the ProAttention mechanism addresses the long-tail phenomenon, reducing the time complexity of the DiT model to $(Llog(L))$. This effectively improves the model's training efficiency and tackles the dimensional challenges posed by large-scale data.
\end{itemize}

\section{Method}

% The entire process of our proposed system is illustrated in Figure 2, which comprises three components: the HRIG module, the TE module, and the EFAttention module. These modules enable us to efficiently generate precise high-resolution fine-grained images. Next, we will provide an introduction to HRIG in Section \ref{HRIG}, followed by an introduction to TE in Section \ref{TE}, and finally, an explanation of the EFAttention module in Section \ref{ESA}.

% We introduce the concept of superclass and divide the denoising process into a semantic generation stage and a perceptual generation stage, incorporating a super-resolution strategy in the perceptual generation stage. Additionally, we propose a ProAttention mechanism to improve training efficiency. Figure \ref{fig:struction} illustrates the architecture of EFDiT.

EFDiT is an improved version based on the DiT architecture and consists of three key modules: High-resolution Image Generation (HRIG), Tiered Embedder (TE), and ESAttention, as shown in Figure \ref{fig:struction}.

% \begin{figure}[!t]\centering
% 	\includegraphics[width=\linewidth]{ICME/figure/Denoising.pdf}
% 	\caption{The denoising process is divided into a semantic information generation phase and a perceptual information generation phase, and the process is regulated using the hyperparameter $\sigma$.}
%     \label{fig:semantic}
% \end{figure}

\subsection{High-resolution Image Generation} \label{HRIG}

Inspired by super-resolution studies \cite{karras2020analyzing}, we introduce an FFT-based enhancement and degradation model into the denoising process of diffusion models. Specifically, we first set the hyperparameter $\sigma$ to transition the model into the perceptual generation stage, where the latent feature map in the current model is represented as: $x_{t} \in \mathbb{R}^{ H \times W \times C}$. To generate clean sampled results that preserve the details required for the perceptual stage, we estimate $x_0$ at each step and denote it as $x_{0|t}$, which can be formulated as:
\begin{equation}
    x_{0|t} = \frac{1}{\sqrt{\bar{\alpha}_{t}}}(x_{t}-\epsilon_{\theta}(x_{t}, t) \sqrt{1- \bar{\alpha}_{t}}),
\end{equation}
where $\epsilon_{\theta} \in \mathbb{R}^{H \times W \times C}$ is the predicted noise value, the prediction of the denoised final image feature map at step $t$ is $x_{0|t} \in \mathbb{R}^{H \times W \times C}$.

Then, we transform the feature map into the frequency domain, preserving its high-frequency information to enhance details and low-frequency information to simulate the degradation process. Finally, we convert the frequency domain features back to a feature map. The formula is as follows:

% Then, during the perception stage, we perform a Fast Fourier Transform (FFT) on the input image $y$ and the sampled result $x_{0|t}$ to obtain their frequency characteristics. High-frequency information typically contains relevant detailed features of the image, while low-frequency information can be used to construct a degradation model, providing targeted and ideal degradation patterns to guide the recovery process, helping the generative model restore finer details and features. Therefore, we obtain the high-frequency and low-frequency information through Gaussian High-pass (GHPF) and Low-pass filters (GLPF), respectively. Next, we obtain the updated high-frequency sampled result $\hat{x}_{{0|t}}^{h}$ and low-frequency sampled result $\hat{x}_{{0|t}}^{l}$ through the Inverse Fast Fourier Transform (IFFT), which are defined as follows:

\begin{equation}
    \hat{x}_{{0|t}}^{h} = \phi^{-1}( \mathbb{G}_{h}(\phi(x_{0|t}))),
\end{equation}

\begin{equation}
    \hat{x}_{{0|t}}^{l} = \phi^{-1}(\mathbb{G}_{l}(\phi(x_{0|t}))),
\end{equation}
where $\hat{x}_{{0|t}}^{h} \in \mathbb{R}^{H \times W \times C}$, $\hat{x}_{{0|t}}^{l} \in \mathbb{R}^{H \times W \times C}$, and $\hat{x}_{{0|t}}^{h}$ represent high-frequency features, and $\hat{x}_{{0|t}}^{l}$ represents low-frequency features. $\phi(\cdot)$ denotes the FFT2d function, $\mathbb{G}_{h} (\cdot)$ denotes the Gaussian High-pass (GHPF), $\mathbb{G}_{l}(\cdot)$ denotes the Gaussian Low-pass filters (GLPF), and $ \phi^{-1}(\cdot) $ denotes the Inverse FFT2d function.

At this point, we preserve the details of the high-frequency information, while also needing to maintain the similarity of the current image $x_{0|t} $ after preserving the high-frequency information. Therefore, we reconstruct the high-frequency loss $\mathcal{L}_{high,pix}$:
\begin{equation}
    \mathcal{L}_{high,pix} = D_{KL}(x_{0|t} (1+\hat{x}_{{0|t}}^{h})|| x_{0|t}),
\end{equation}
At this point, we supplement the low-frequency information $ \hat{x}_{{0|t}}^{l} $ from the degradation model into the enhanced model $x_{0|t}  (1+\hat{x}_{{0|t}}^{h}) \in \mathbb{R}^{ H \times W \times C}$, resulting in a new sampled output $ \hat{x}_{{0|t}}^{a} \in \mathbb{R}^{ H \times W \times C}$. The next state $x_{t-1}$ can be sampled from a joint distribution, which is formulated as:
\begin{equation}
    p_{\theta}(x_{t-1}|x_{t}, \hat{x}_{{0|t}}^{a}) = \mathcal{N}(x_{t-1};\mu_{\theta}(x_{t}, \hat{x}_{{0|t}}^{a}), \sigma^2_{t} \mathbf{I}),
\end{equation}
where $\mu_{\theta}(x_{t},\hat{x}_{{0|t}}^{a}) = \frac{\sqrt{\bar{\alpha}_{t}-1}\beta_{t}}{1-\bar{\alpha}_{t}}\hat{x}_{{0|t}}^{a} + \frac{\sqrt{\alpha_{t}}(1-\bar{\alpha}_{t-1})}{1-\bar{a}_{t}}x_{t}$ and $\sigma^2_{t} = \frac{1- \bar{\alpha}_{t-1}}{1-\bar{\alpha}_{t}}\beta_{t}$. By using the FFT-based enhancement and degradation model to guide the sampling in the perceptual generation stage, we generate the result $ x_0 \in \mathbb{R}^{ H \times W \times C} $ with image detail features.

Although the generated result $ x_{0} $ has richer details compared to before, we still aim to explore the specific proportions of the degradation model and enhancement model that control these details. Therefore, we introduce a scientifically-based adaptive factor $\gamma$ to control the proportion of the enhancement and degradation models during the sampling process. Additionally, we introduce a reconstruction loss $\mathcal{L}_{rec}$  to ensure that $ x_{0|t} $ remains as close as possible to $ x_{0} $ overall:
\begin{equation}
     \mathcal{L}_{rec} = D_{KL}( ( \gamma (x_{0|t}(1+\hat{x}_{{0|t}}^{h}) + (1-\gamma)\hat{x}_{{0|t}}^{l}) || x_{0|t}). 
\end{equation}

Considering all the aforementioned losses, the overall loss is formalized as:

\begin{equation} 
\mathcal{L}_{EFD} = \mathcal{L}_{org} + \lambda_{1}\mathcal{L}_{hig,pix} + \lambda_{2}\mathcal{L}_{rec}, 
\end{equation}
where \( L_{\text{org}} \) is the original loss function in the diffusion model. For simplicity, we use $\lambda_{1}=\lambda_{2}=1$ in this paper.

% \begin{figure}[!t]\centering
% 	\includegraphics[width=\linewidth]{ICME/figure/FFT_2.pdf}
% 	\caption{Image Enhancement Pipeline. The edges and contours of fine-grained images are enhanced using FFT and Gaussian filtering, thereby improving the relevant details in the generated fine-grained images.}
%     \label{fig:enh}
% \end{figure}

\subsection{Tiered Embedder} \label{TE}

In the embedding layer, accurately embedding category information is crucial for generating precise category images, especially in fine-grained datasets. However, existing fine-grained image generation models typically only embed the corresponding subclass labels. To address this issue, this paper introduces a “tiered embedder" strategy. The approach attempts to better integrate superclass and subclass information while ensuring that semantic information is fed into the diffusion model for training. For this, we assume the subclass label is $ c_{s1} \in \mathbb{R}^{d_1}$ and the superclass label is $ c_{s2} \in \mathbb{R}^{d_2}$. In a conditional diffusion model $\epsilon_{\theta}(x_{t}|c) \in \mathbb{R}^{ H \times W \times C}$, $c$ is the condition and $ x_{t} \in \mathbb{R}^{ H \times W \times C} $ is the noisy sample. The predicted value $ \hat{\epsilon}(x_{t}|c) \in \mathbb{R}^{ H \times W \times C} $ is given by:
\begin{equation}
    \hat{\epsilon}(x_{t}|c) = \epsilon_{\theta} 
      (x_{t}|\emptyset) + \omega (\epsilon_{\theta}(x_{t}|c) - \epsilon_{\theta}(x_{t}|\emptyset)),
\end{equation}
where $\omega \geq  1$ is the guidance scale. 

% We incorporate the subclass label information $c_{s1}$ and the superclass label information $c_{s2}$ as follows:
% \begin{equation}
%     \hat{\epsilon}(x_{t}|\{c_{s1},c_{s2}\}) = \epsilon_{\theta} 
%       (x_{t}|\emptyset) + \omega (\epsilon_{\theta}(x_{t}|c) - \epsilon_{\theta}(x_{t}|\emptyset))
% \end{equation}
We integrate the superclass and subclass conditional diffusion models $\epsilon_{\theta}(x_{t}|\{c_{s1}, c_{s2}\}$ into $\epsilon_{\theta}(x_{t}|\{c_{si}\}_{i=1}^{i=2})$, where $c_{s2}$ are the superclass label conditions and $c_{s1}$ are the subclass label conditions. During inference, we calculate the direction for each condition $\Delta_{i}^{t} = \epsilon_{\theta}(x_{t}|c_{si}) - \epsilon_{\theta}(x_{t}|\emptyset) $, and linearly combine them using two guidance scales $\omega_{i}$:

\begin{equation}
    \hat{\epsilon}(x_{t}|c) =  \epsilon_{\theta} +  \sum_{i=1}^{i=2} \omega_{i} \Delta_{i}^{t},
\end{equation}
for convenience, we set \(\omega_1\) and \(\omega_2\) to 1 in this paper.

With the above formulation, we are able to fully embed the label information of both the superclass and subclass from fine-grained image datasets, and separately control the corresponding conditions during inference. Subsequently, by fine-tuning the bias and normalization layers after the embedding layer (as shown in Figure \ref{fig:struction}), the model can better learn the data distribution.

\subsection{Efficient Self-attention Mechanism} \label{ESA}
The attention mechanism is based on tuple input, defined by $q$, $k$, $v$, and its defined formula is: $A(Q,K,V)=Softmax(\frac{QK^{T}}{\sqrt{d}})V$
Where d is the input dimension. In order to further analyze the attention mechanism, this article represents $q_i$,$k_i$,$v_i$ as the i-th row of $Q$, $K$, $V$. The probability form of the $i-th$ query attention can be defined as: 

\begin{equation} 
A(q_{i},K,V)=\sum \limits_{j}\frac{k(q_{i},k_{i})}{\sum_{l} k(q_{i},k_{l})}v_{j}= \mathbb{E}_{p(k_{j}|q_{i})} \left[ v_{j} \right] ,
\end{equation}
when the $i$-th query gets a larger $M(q_{i},K)$, it has a higher chance of containing the dominant dot product pair in the header field of this long-tail self-attention distribution. With the proposed formula, we only need to focus on the u dominant queries by each $k$:
\begin{equation} A(Q,K,V)=Softmax(\frac{\Bar{Q}K^{T}}{\sqrt{d}})V, 
\end{equation}
$\Bar{Q}$ is a sparse matrix of the same size as q, which contains the top-u query in the sparsity of two $M(q, K)$. 

When we set a constant sampling factor $c$, we set u = $c· InL_Q$ For each self-attention q, only the dot product of $O(In L_{Q})$ needs to be calculated. For each query $q_{i}\in{\mathbb{R}^{d}}$ and $k_{j} \in{\mathbb{R}^d}$ in the set $\bm{K}$, we have the bound as: $InL_{K} \leq M(q_{i},K) \leq max_{j} \{q_{i}k_{j}^T/\sqrt{d} \} - \frac{1}{L_{K}}\sum_{j=1}^{L_{K}}\{q_{i}k_{i}^{T}/\sqrt{d}\}+InL_{K}$. 
When $q_{i}\in K$
, it also holds. 
So we propose the max-mean measurement as: 
\begin{equation}
\bar{M}(q_{i},K) = max_{j}\{ \frac{q_{i}k_{j}^{T}}{\sqrt{d}}\} - \frac{1}{L_{K}}\sum\limits^{L_{K}}_{j=1}\frac{q_{i}k_{j}^{T}}{\sqrt{d}}, \end{equation}
under the long tail distribution, we need to randomly sample $U = L_{K}InL_{Q}$ dot-product pairs to calculate the $\Bar{M}(q_{i},K)$. 

Then we select Top-u from $\bar{Q}$. Actually, In self-attention computing, the input lengths of queries and keys are usually equivalent, i.e. $L_Q = L_K = L$. Therefore, the proattention sub-attention time complexity is $O(LInL)$ (The detailed derivation process can be found in the appendix).

\section{Experiment}
% In our experiments, we used the iNaturalist 2021 dataset and the VegFru dataset as the training datasets for our model. Both datasets are fine-grained, with the iNaturalist 2021 dataset containing 10,000 different species across 11 supercategories. Compared to the ImageNet-1k dataset, the iNaturalist dataset has a richer set of classification units, making it more challenging to generate similar species. The VegFru dataset comprises 292 fine-grained categories of vegetables and fruits, organized into 25 superclasses. The pre-trained model we utilize is a DiT XL/2 model as our backbone. 
In our experiments, we used the iNaturalist 2021 dataset and the VegFru dataset as the training datasets for our model. Both datasets are fine-grained, with the iNaturalist 2021 dataset containing 10,000 different species across 11 supercategories. The VegFru dataset comprises 292 fine-grained categories of vegetables and fruits, organized into 25 superclasses. The pre-trained model we utilize is a DiT XL/2 model as our backbone.

\begin{figure}[!t]\centering
	\includegraphics[width=\linewidth]{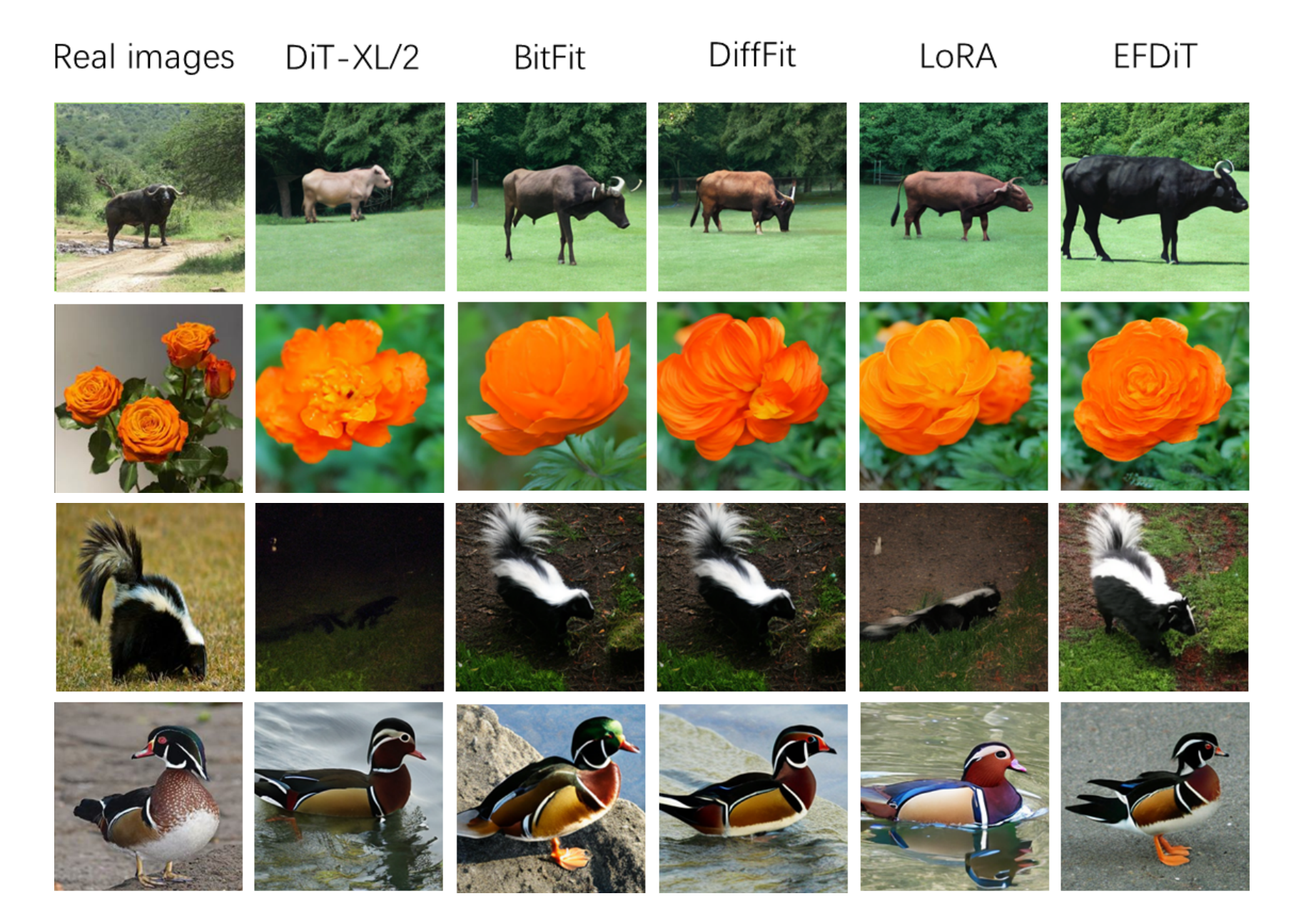}
	\caption{Comparison with fine-grained images generated by other algorithms.}
    \label{pic:compare}
\end{figure}

\subsection{Implementation Details}
% To generate high-resolution fine-grained images, we employed a two-stage training method. In the first stage, we introduced the concept of superclasses and fine-tuned the pre-trained model to adapt to the more diverse labels in the iNaturalist 2021 \cite{van2018inaturalist_intro} and VegFru \cite{hou2017vegfru} fine-grained datasets. The fine-tuning was completed over 200k iterations, with weights of $\{\alpha,\beta \}$ set to $\{0.8,0.8 \}$. In the second stage, we incorporated the concept of super-resolution, introducing a new KL divergence loss during the perceptual image generation phase, with the weights of $\gamma$ set to $0.7$. We used the Adam optimizer with a learning rate of 1e-4. The entire training process was conducted on an Nvidia V100. For the iNaturalist 2021 fine-grained dataset, the image resolution was 256×256, with a batch size of 2. The sampling process adopted classifier-free guidance with a scale of 4.0 and used 250 steps for sampling. Similarly, for the VegFru fine-grained dataset, the image resolution was 256×256, with a batch size of 2, and the sampling process also adopted classifier-free guidance with a scale of 4.0 and 100 steps for sampling.  

We utilize the pretrained DiT-XL/2 as the backbone. The fine-tuning is completed over 200k iterations. We use the Adam optimizer with a learning rate of 1e-4 and set $\gamma$ to 0.7. The entire training process is conducted on an Nvidia V100. For the iNaturalist 2021 fine-grained dataset, the image resolution is 256×256, with a batch size of 2. The sampling process adopts classifier-free guidance with a scale of 4.0 and uses 250 steps for sampling. Similarly, for the VegFru fine-grained dataset, the image resolution is 256×256, with a batch size of 2, and the sampling process also adopts classifier-free guidance with a scale of 4.0 and 100 steps for sampling.  
% \begin{figure}[!t]\centering
% 	\includegraphics[width=\linewidth]{sec/figure/sd_fish.pdf}
% 	\caption{Comparison with stable diffusion in the generation of various fine-grained fish categories.}
%     \label{sd_fish}
% \end{figure}

\begin{table*}[t]
\caption{FID and IS metrics comparison with existing methods DiT-XL/2 (Full Fine-tuning)~\cite{peebles2023scalable}, BitFit~\cite{zaken2021bitfit}, DiffFit~\cite{xie2023difffit}, LoRA~\cite{hu2021lora} and DiffiT~\cite{hatamizadeh2025diffit} on 11 categories of iNaturalist 2021 mini dataset image synthesis. }
\label{tab:metrics_comparison}
\centering
\renewcommand\arraystretch{1.3}
\setlength{\tabcolsep}{0.09cm}

% Please add the following required packages to your document preamble:
% \usepackage{multirow}
% \usepackage[table,xcdraw]{xcolor}
% If you use beamer only pass "xcolor=table" option, i.e. \documentclass[xcolor=table]{beamer}
\scalebox{0.96}{
\begin{tabular}{ccccccccccccc cccccccccccc}
\cline{1-25}
                         & \multicolumn{2}{c}{Plants}            & \multicolumn{2}{c}{Insects}       & \multicolumn{2}{c}{Birds}         & \multicolumn{2}{c}{Fungi}         & \multicolumn{2}{c}{Reptiles}      & \multicolumn{2}{c}{Mammals}         & \multicolumn{2}{c}{Fishes} & \multicolumn{2}{c}{Amphibians}    & \multicolumn{2}{c}{Mollusks}      & \multicolumn{2}{c}{Arachnids}     & \multicolumn{2}{c}{Animalia}      & \multicolumn{2}{c}{Overall} \\ \cline{2-25}
\multirow{-2}{*}{Method}  & FID$\downarrow$   & IS$\uparrow$   & FID$\downarrow$ & IS$\uparrow$ & FID$\downarrow$ & IS$\uparrow$ & FID$\downarrow$ & IS$\uparrow$ & FID$\downarrow$ & IS$\uparrow$ & FID$\downarrow$ & IS$\uparrow$  & FID$\downarrow$   & IS$\uparrow$   & FID$\downarrow$ & IS$\uparrow$ & FID$\downarrow$ & IS$\uparrow$ & FID$\downarrow$ & IS$\uparrow$ & FID$\downarrow$ & IS$\uparrow$ & FID$\downarrow$ & IS$\uparrow$  \\ \cline{1-25}
DiT-XL/2~\cite{peebles2023scalable}          & 13.2           & 5.8             & 18.4          & 5.6           & 21.5          & 6.4           & \textbf{27.2}          & 3.2           & 37.8          & 4.7        & 38.1          & 7.4         & 35.6            & 4.2             & 55.1          & 2.1           & 44.1          & 3.7           & 45.7          & 3.1           & 50.0          & 3.9           & 10.7          & 11.4     \\
BitFit~\cite{zaken2021bitfit}                   & \textbf{12.0}            & 6.3             & 19.3          & 5.9           & 23.6          & 6.4           & 29.5          & \textbf{5.6}           & 41.8          & 4.6           & 39.9          &  7.4        & 41.2            & 4.2             & 66.2          & 1.8           & 44.5          & 3.9           & 45.2          & 2.9           & 51.9          & 4.2           & 10.5          & 11.7      \\
DiffFit~\cite{xie2023difffit}                   & 12.1                     & 6.2             & 19.8          & 5.8           & 24.0          & 6.3           & 30.0          & 3.3           & 42.6          & 4.5           & 40.2          & 7.5         & 44.3            & 4.1             & 68.0          & 1.9           & 44.8          & 3.9           & 45.8          & 3.1           & 52.1          & 4.3           & 10.7          & 11.8          \\
LoRA~\cite{hu2021lora}                   & 12.1                         & 6.1             & 18.4          & 5.9           & 22.5          & 6.3           & 28.4          & 3.3           & 40.0          & 4.6           & 39.1          & 7.7           & 40.9            & 3.9             & 62.3          & 1.9           & 43.5          & 4.1           & 46.2          & 3.1           & 54.9          & 4.2           & 10.2          & 11.6     \\
DiffiT~\cite{hatamizadeh2025diffit}                   & 12.4                         & 6.3             & 18.2          & 5.9           & 22.3      &    6.4         & 28.3          & 4.8           & 37.7          & 5.0          & 38.1          & 7.9         & 34.2            & 4.3             & 48.5          & 2.1           & 40.7          & 5.0          & 40.7          & 3.6           & 50.0          & 4.3           & 10.1          & 11.9       \\
\rowcolor{gray!20}
EFDiT\textbf{(ours)}      & 12.5            & \textbf{7.1}            & \textbf{14.0}          & \textbf{6.6}           & \textbf{14.7}          & \textbf{7.3}           & 27.7          & 5.1           & \textbf{25.2}          & \textbf{5.7}           & \textbf{33.8}          & \textbf{8.4}        &  \textbf{30.2}            & \textbf{5.5}                     &     \textbf{45.8}         & \textbf{3.3}           & \textbf{37.9}          & \textbf{5.2}           & \textbf{36.5}          & \textbf{4.3}           & \textbf{48.9}           & \textbf{4.9}           & \textbf{9.6}  & \textbf{12.6}     \\ \cline{1-25}
\end{tabular}
}

\label{tab:main_fid_IS_result}
\end{table*}

\subsection{Comparisons with State-of-the-art Methods}

We quantitatively and qualitatively compare our EFDiT with other SOTA fine-tuning and common methods on fine-grained dataset generation tasks, including DiT-XL/2 (Full Fine-tuning) \cite{peebles2023scalable}, BitFit \cite{zaken2021bitfit}, DiffFit \cite{xie2023difffit}, LoRA \cite{hu2021lora} and DiffiT \cite{hatamizadeh2025diffit}.

\begin{table}[!t]
\caption{Overall FID and IS metrics on the VegFru dataset, and number of parameters and training overhead on the iNaturalist dataset.}\label{tab:vegfru_result}
\centering
\renewcommand\arraystretch{1}
\setlength{\tabcolsep}{0.20cm}
\scalebox{0.97}{
\begin{tabular}{ccccc cc}
\hline
Method            & FID$\downarrow$      & IS $\uparrow$  & Params.(M)$\downarrow$    & Time.(D)$\downarrow$ & Speed$\downarrow$ \\ \hline
DiT-XL/2~\cite{peebles2023scalable}     & 13.034         & 8.505       & 685.2(100\%)  & 15.11          & 1$\times$                                                                                                       \\
BitFit~\cite{zaken2021bitfit}              & 15.022            & 9.487        & 12.01(1.75\%) & 9.59                                                               & 0.635$\times$                                                                                               \\
DiffFit~\cite{xie2023difffit}           & 15.068              & 9.397      & 12.14(1.77\%) & 9.65                                                               & 0.638$\times$                                                                                                                               \\
LoRA~\cite{hu2021lora}               & 14.549                 & 9.670      & 12.56(1.83\%) & 9.92  & 0.657$\times$                                         
                                            \\
DiffiT~\cite{hatamizadeh2025diffit}              &    13.528         &9.593       & 561.24(100\%)  & 14.46  & 1$\times$     \\
% \rowcolor{gray!20}
EFDiT\textbf{(ours)}  & \textbf{12.425}               & \textbf{9.712}     &    \textbf{11.52(1.68\%)}   & \textbf{9.51}                                                      &         \textbf{0.629$\times$}                                                                                     \\ \hline
\end{tabular}}
% \vspace{-5mm}
\end{table}

\subsubsection{Quantitative Comparison}

We conducted training and inference on the fine-grained datasets iNaturalist 2021 and VegFru, and evaluated them using the FID \cite{heusel2017gans} and IS \cite{salimans2016improved} metrics. The FID metric reflects the similarity between real and generated images through trace values. The IS metric evaluates the quality of generated images from two aspects: the clarity of individual generated images and the diversity of the generated images. For fine-grained image generation models, the diversity of the generated fine-grained images is also an important indicator of the model's performance. The clearer and more diverse the generated images are, the higher the IS score.

We trained on fine-grained datasets and compared the FID and IS metrics for each category's generated images. As shown in Table~\ref{tab:metrics_comparison} and~\ref{tab:vegfru_result}, we compared the performance of EFDiT with existing methods in category-conditioned fine-grained image generation. In the iNaturalist 2021 fine-grained dataset, Table~\ref{tab:metrics_comparison} presents the FID and IS metrics for 11 fine-grained supercategories as well as overall metrics. It can be observed that the DiffFiT model does not perform well in terms of image generation quality on the iNaturalist 2021 fine-grained dataset. For example, in categories such as Insects (19.8, 18.4), Birds (24, 21.5), Mammals (40.2, 38.1), and Fishes (44.3, 35.6), its FID metrics are worse than its Backbone DiT. The model was fine-tuned on DiT by adjusting Bias and Norm among other components. We found that merely fine-tuning these modules does not adapt well to fine-grained datasets because, compared to the ImageNet-1k dataset, fine-grained datasets have more categories with higher similarity between them. The proposed model in this paper incorporates a ``tiered embedder" model and a super-resolution concept module on top of fine-tuning, which enables it to generate higher quality datasets in the iNaturalist 2021 fine-grained dataset. For instance, in categories like Insects (14.0, 18.4), Birds (14.7, 21.5), Mammals (33.8, 38.1), and Fishes (30.2, 35.6), the
FID metrics are better than those of the DiT model.

For fine-grained datasets, the diversity of generated images is also an important metric. Therefore, this paper uses the IS metric to evaluate the model's results. The findings reveal that almost all models fine-tuned based on the DiT model have IS metrics for generating fine-grained images that are better than the DiT model itself. For example, the DiffFit model shows superior IS metrics in categories such as Insects (5.8, 5.6), Mammals (7.5, 7.4), and Animalis (4.3, 3.9) compared to DiT, indicating that fine-tuning the DiT model is beneficial for enhancing the diversity of fine-grained image generation. The model proposed in this paper incorporates the concept of super-resolution and a “tiered embedder" model, which results in FID and IS metrics that are superior to other compared models across most categories. Additionally, by fine-tuning only 1.68\% of the parameters, the overall FID on the iNaturalist dataset reaches 9.638. As shown in Table~\ref{tab:vegfru_result}, for the VegFru dataset, it is observed that the FID metrics of the fine-tuned models are not as good as those of the DiT model. However, for the IS metric, BitFit (9.487, 8.595), DiffFit (9.397, 8.595), LoRA (9.670, 8.595), DiffIT (9.593, 8.595), and EFDiT (9.712, 8.595) all outperform the DiT model, indicating that fine-tuning is beneficial for enhancing the diversity of fine-grained image generation. EFDiT achieves state-of-the-art FID and IS metrics on both the iNaturalist and VegFru datasets by fine-tuning only 1.68\% of the parameters. The results demonstrate that the modules proposed in this paper can generate high-quality, diverse fine-grained images.

\begin{figure}[t]\centering
	\includegraphics[width=\linewidth]{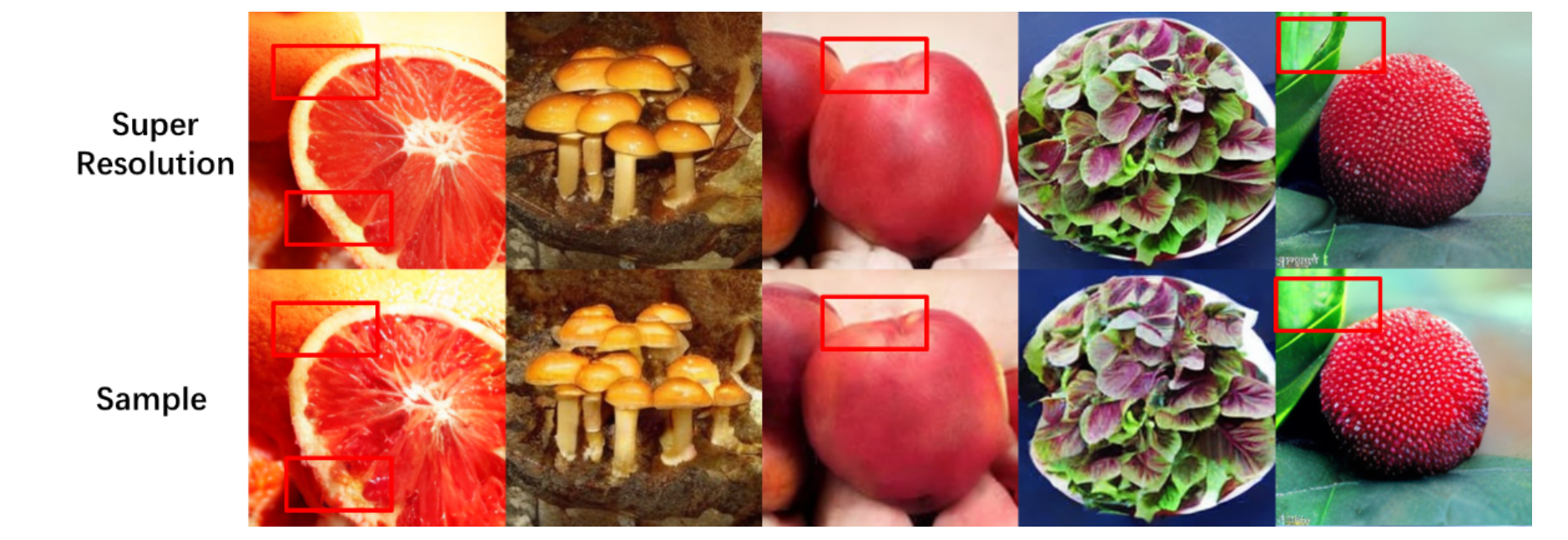}
	\caption{Comparison between super-resolution and sampling in image generation.}
    \label{pic:super_resolution}
\end{figure}

\begin{figure}[!t]\centering
	\includegraphics[width=\linewidth]{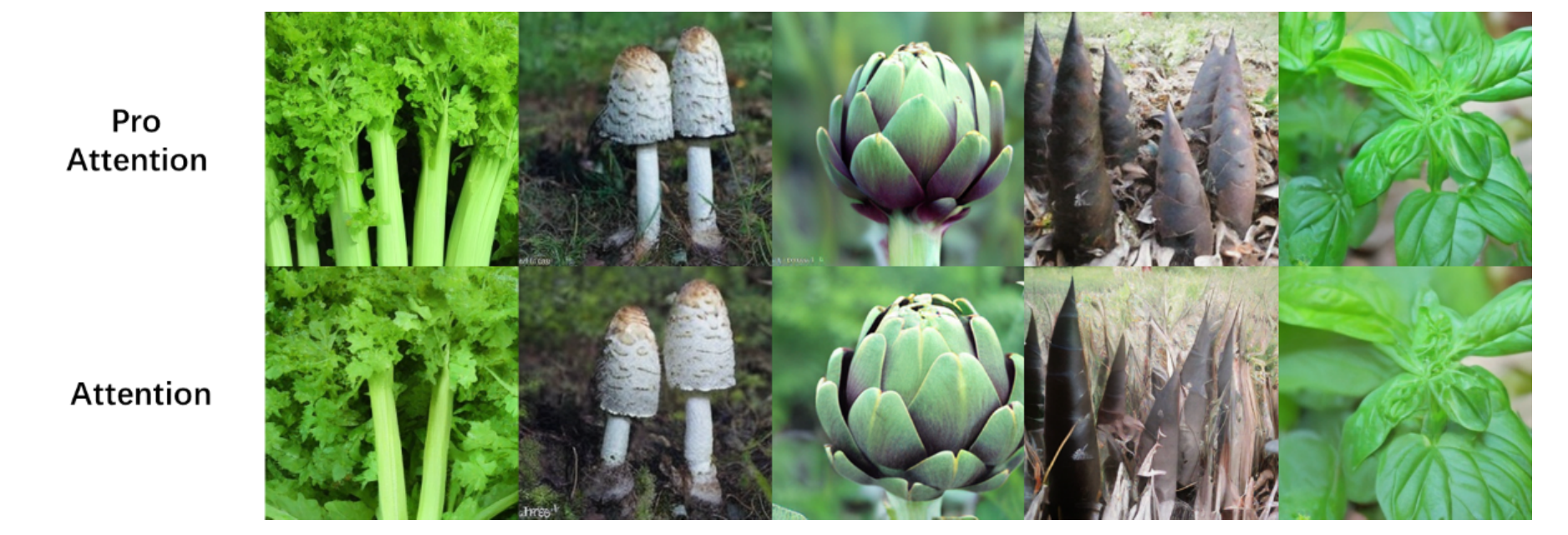}
	\caption{Comparison between the ProAttention mechanism and Attention mechanism in image generation.}
    \label{pic:pro_attention}
\vspace{-2em}
\end{figure}

\subsubsection{Qualitative Comparison}
As shown in Figure \ref{fig:fig1}, high-resolution fine-grained images are generated for similar classes.  As shown in Figure \ref{pic:compare}, EFDiT retains more semantic information during the image generation stage compared to other methods, resulting in more accurate class labels for the generated fine-grained images. By introducing the concept of super-resolution, our model significantly improves the detail features of the generated images, thereby enhancing the quality of the images. For the label-to-image task involving fine-grained images, the EFDiT model proposed in this paper is able to more thoroughly understand the semantic information of fine-grained image labels.

% \begin{table}[t]
% \centering
% % \renewcommand\arraystretch{1.3}
% \renewcommand\arraystretch{1}
% \includegraphics[width=1\linewidth]
% % \setlength{\tabcolsep}{0.055cm}
% % \setlength{\tabcolsep}{0.25cm}
% \begin{tabular}{ccccc}
% \hline
% Method            & FID$\downarrow$      & IS $\uparrow$\\ \hline
% LableEmbedder(DiT)     & 13.034                                                                                                                    \\
% TieredEmbedder     & 15.068 \\
% TieredEmbedder + bias             & 15.068             \\
% TieredEmbedder + norm                & 14.549                                                 
%                                             \\
% % \rowcolor{gray!20}
% EFDiT\textbf{(ours)} & \textbf{12.425}                                                                                               \\ \hline
% \end{tabular}
% \caption{Comparison of the overall FID and overall LPIPS metrics on VegFru dataset.}\label{tab:vegfru_result}
% % \vspace{-5mm}
% \end{table}

\begin{table}[t]
\caption{Comparison of the overall FID and overall IS metrics on VegFru dataset.}\label{tab:qualitative_1s}
\centering
\renewcommand\arraystretch{1}
\setlength{\tabcolsep}{0.25cm}
\begin{tabular}{ccccc}
\hline
Method            & FID$\downarrow$      & IS $\uparrow$\\ \hline
LabelEmbedder(DiT)     & 13.034          & 8.505                                                                                                           \\
TieredEmbedder     & 12.653                  &     9.454                                                                                      \\
TieredEmbedder + bias             & 12.583        &  9.553                                                                                          \\
TieredEmbedder + norm                & 12.604    & 9.512  \\
% \rowcolor{gray!20}
EFDiT\textbf{(ours)} & 12.425     & 9.712                                                                                              \\ \hline
\end{tabular}

% \vspace{-5mm}
\end{table}

% \begin{table}[t]
% \centering
% % \renewcommand\arraystretch{1.3}
% \renewcommand\arraystretch{1}
% % \includegraphics[width=1\linewidth]
% % \setlength{\tabcolsep}{0.055cm}
% \setlength{\tabcolsep}{0.25cm}
% \begin{tabular}{ccccc}
% \hline
% Method            & FID$\downarrow$      & IS $\uparrow$\\ \hline
% LabelEmbedder(DiT)     & 13.034                                                                                                                     \\
% TieredEmbedder(\alpha, \beta)             & 15.022                                                                                                             \\
% TieredEmbedder(\alpha, \beta)              & 15.068                                                                                           \\
% % \rowcolor{gray!20}
% EFDiT\textbf{(ours)} & \textbf{12.425}                                                                                                   \\ \hline
% \end{tabular}
% \caption{Comparison of the overall FID and overall LPIPS metrics on VegFru dataset.}\label{tab:qualitative_1}
% % \vspace{-5mm}
% \end{table}

\begin{table}[!t]
\caption{Comparison of the overall FID and overall IS metrics on VegFru dataset.}
\centering
\renewcommand\arraystretch{1}
\setlength{\tabcolsep}{0.25cm}
\begin{tabular}{cccc}
\hline
Method            & FID$\downarrow$      & IS $\uparrow$\\ \hline
LableEmbedder(DiT)     & 13.034                 & 8.505                                                                                                  \\
EFDiT(Pro)                & 12.851       & 9.102                                               
                                            \\
% \rowcolor{gray!20}
EFDiT\textbf{(ours)} & 12.425     & 9.712                                                                                          \\ \hline
\end{tabular}
\label{tab:qualitative_2}
 \vspace{-2mm}
\end{table}

% \begin{table}[t]
% \caption{Comparison of the overall FID and overall IS metrics on VegFru dataset.}
% \centering
% % \renewcommand\arraystretch{1.3}
% \renewcommand\arraystretch{1}
% % \includegraphics[width=1\linewidth]
% % \setlength{\tabcolsep}{0.055cm}
% \setlength{\tabcolsep}{0.25cm}
% \begin{tabular}{cccc}
% \hline
% Method            & FID$\downarrow$      & IS $\uparrow$\\ \hline
% EFDiffusion     & 13.034                                                                                                                 \\
% EFDiT(\gamma)             & 15.068                                                                                                               \\
% EFDiT(\gamma)                & 14.549                                                                 
%                                             \\
% % \rowcolor{gray!20}
% EFDiT\textbf{(ours)} & \textbf{12.425}                                                                                                      \\ \hline
% \end{tabular}
% \label{tab:qualitative_3}
% % \vspace{-5mm}
% \end{table}

\subsection{Ablation Study}
In this section, we conduct ablation experiments to assess the substantive impact of introducing superclasses, ProAttention, and super-resolution on generating high-quality fine-grained images in fine-grained image datasets. 

Table \ref{tab:qualitative_1s} evaluates the fine-tuning of the model with the introduction of a tiered embedder on the Vegfru dataset for 256×256 fine-grained images using FID and IS metrics. The results show that the introduction of the tiered embedder significantly improves the IS metric compared to the model without it, indicating a notable enhancement in the diversity of fine-grained image generation. Table \ref{tab:qualitative_2} evaluates the introduction of Pro-Attention and the super-resolution model on the Vegfru dataset using FID and IS metrics. The results show that the introduction of the super-resolution model has a greater impact on the FID metric compared to Pro-Attention.

% Figure 6 compares the FID and IS scores of the DiT model, the model with only ProAttention, and the EFDiT model for generating 256×256 fine-grained images on the Inaturalist dataset.

% In the ablation experiments, we evaluate the impact of introducing superclasses, ProAttention, and super-resolution on fine-grained image generation. Visual comparisons reveal that ProAttention, as shown in Figure \ref{pic:pro_attention}, maintains image quality comparable to standard attention while reducing time and space complexity. Figure \ref{pic:super_resolution} highlights the difference between fine-grained images generated with and without the super-resolution concept. The red matrix indicates that the super-resolution concept effectively preserves contours at different pixel boundaries, enhancing the quality of fine-grained image generation.

In the ablation experiments, we conducted a qualitative comparison of the model to evaluate the impact of ProAttention and super-resolution on fine-grained image generation. As shown in Figure \ref{pic:pro_attention}, ProAttention maintains relatively better image quality compared to standard attention. Figure \ref{pic:super_resolution} highlights the difference in the quality of fine-grained image generation with the introduction of the super-resolution model. As indicated by the red matrix, the super-resolution concept effectively preserves contours at different pixel boundaries, enhancing the quality of fine-grained image generation.

\section{Conclusion}
% In this paper, we thoroughly investigate the operational principles of diffusion models and utilize their features to achieve high-resolution generation of fine-grained image datasets through efficient fine-tuning strategies. Specifically, we introduce the concept of superclasses, which allows the model to retain sufficient semantic information in fine-grained image datasets where semantic information is relatively sparse, thereby accurately guiding the generation of fine-grained categories. Upon discovering a long-tail phenomenon in image generation with the DiT model, we introduce a ProAttention mechanism that reduces the time and space complexity of the DiT model to $Llog(L)$, enabling more efficient training and inference, which is crucial for models constrained by image pixel size. Additionally, during the perceptual information generation phase of the diffusion model, we introduce a novel super-resolution concept using an image degradation model and an image contour enhancement strategy, allowing the diffusion model to generate high-quality fine-grained images. Extensive experimental results show that our method surpasses existing SOTA methods. By incorporating innovative fine-tuning strategies, we not only efficiently generate high-quality fine-grained super-resolution images on fine-grained image datasets but also significantly enhance the accuracy and visual quality of the generated images.
In this paper, we thoroughly investigate the operational principles of diffusion models and utilize their features to achieve high-resolution generation of fine-grained image datasets through efficient fine-tuning strategies. Specifically, we introduce the “tiered embedder" module and the concept of super-resolution, which enable the model to retain sufficient semantic information in fine-grained image datasets where semantic information is relatively sparse, thereby accurately guiding the generation of high-quality fine-grained images. Additionally, we introduce an efficient self-attention mechanism to enable more efficient training and inference. Extensive experimental results demonstrate that our method surpasses existing state-of-the-art methods.

In future work, we will evaluate the impact of the fine-grained dataset itself on the model. The EFDiT proposed in this paper has shown significant improvements in image quality and category diversity for specific categories compared to other models, but there are still limitations in improving overall metrics. This phenomenon is quite interesting, and we initially speculate that it may be related to the long-tail nature of the dataset. Therefore, we will continue to analyze the model structure and related data and aim to develop models capable of generating accurate, high-quality fine-grained images.
% 决定方法存在的问题,为什么没解决.

\section{ACKNOWLEDGEMENT}
% This work was supported by the National Key Re
% search and Development Program of China (Grant No. 2020AAA0108003) and National Natural Science Foundation
%  of China (GrantNo.62277011).

 This work was supported by the National Key Research and Development Program of China (Grant No. 2020AAA0108003), National Natural Science Foundation of China (Grant No. 62277011) and Project of Chongqing MEITC(Grant No. YJX-2025001001009)
% In this paper, we delve into the operational mechanisms of diffusion models and achieve high-resolution generation of fine-grained image datasets through efficient fine-tuning strategies. We introduce the concept of superclasses to address the semantic entanglement issues in fine-grained image datasets, thereby accurately guiding the generation of fine-grained categories. Upon identifying a long-tail phenomenon in image generation with the DiT model, we propose a ProAttention mechanism that reduces the time and space complexity of the DiT model to $L\log(L)$, enabling more efficient training and inference. Additionally, during the perceptual information generation phase of the diffusion model, we introduce an innovative super-resolution concept that combines an image degradation model with an image contour enhancement strategy, allowing the diffusion model to produce high-quality fine-grained images. Extensive experimental results demonstrate that our method not only efficiently generates high-quality fine-grained super-resolution images on fine-grained image datasets but also significantly enhances the accuracy and visual quality of the generated images.

% \section*{Acknowledgment}

% The preferred spelling of the word ``acknowledgment'' in America is without 
% an ``e'' after the ``g''. Avoid the stilted expression ``one of us (R. B. 
% G.) thanks $\ldots$''. Instead, try ``R. B. G. thanks$\ldots$''. Put sponsor 
% acknowledgments in the unnumbered footnote on the first page.

\bibliographystyle{IEEEbib}
\bibliography{arxiv}

% \vspace{12pt}
% \color{red}
% IEEE conference templates contain guidance text for composing and formatting conference papers. Please ensure that all template text is removed from your conference paper prior to submission to the conference. Failure to remove the template text from your paper may result in your paper not being published.

\end{document}